\documentclass[sigconf]{acmart}

\DeclareMathOperator*{\argmin}{\arg\!\min}

\usepackage{mathtools}
\DeclarePairedDelimiterX{\infdivx}[2]{(}{)}{%
  #1\;\delimsize\|\;#2%
}

\newcommand{\kld}{D_{KL}\infdivx}

\acmConference[MILETS: SIGKDD International Workshop on Mining and Learning from Time Series]{MILETS: SIGKDD International Workshop on Mining and Learning from Time Series -- Deep Forecasting: Models, Interpretability, and Applications}{2022}{Washington DC, USA}

\acmYear{}
\acmPrice{}
\acmISBN{}
\acmDOI{}
\setcopyright{none}

\begin{document}

\title{Domain Adaptation Under Behavioral and Temporal Shifts for Natural Time Series Mobile Activity Recognition}

 \author{Garrett Wilson}
 \email{garrett.wilson@wsu.edu}
 \orcid{0000-0002-6760-754X}
 \affiliation{%
   \institution{Washington State University}
   \streetaddress{School of Electrical Engineering and Computer Science}
   \city{Pullman}
   \state{WA}
   \country{USA}
   \postcode{99164}
 }

 \author{Janardhan Rao Doppa}
 \email{jana.doppa@wsu.edu}
 \orcid{0000-0002-3848-5301}
 \affiliation{%
   \institution{Washington State University}
   \streetaddress{School of Electrical Engineering and Computer Science}
   \city{Pullman}
   \state{WA}
   \country{USA}
   \postcode{99164}
 }

 \author{Diane J. Cook}
 \email{djcook@wsu.edu}
 \orcid{0000-0002-4441-7508}
 \affiliation{%
   \institution{Washington State University}
   \streetaddress{School of Electrical Engineering and Computer Science}
   \city{Pullman}
   \state{WA}
   \country{USA}
   \postcode{99164}
 }

\renewcommand{\shortauthors}{Wilson, Doppa, and Cook}
\renewcommand{\shorttitle}{Domain Adaptation Under Behavioral and Temporal Shifts}

\begin{abstract}
Increasingly, human behavior is captured on mobile devices, leading to an increased interest in automated human activity recognition. However, existing datasets typically consist of scripted movements. Our long-term goal is to perform mobile activity recognition in natural settings. We collect a dataset to support this goal with activity categories that are relevant for downstream tasks such as health monitoring and intervention. Because of the large variations present in human behavior, we collect data from many participants across two different age groups. Because human behavior can change over time, we also collect data from participants over a month’s time to capture the temporal drift. We hypothesize that mobile activity recognition can benefit from unsupervised domain adaptation algorithms. To address this need and test this hypothesis, we analyze the performance of domain adaptation across people and across time. We then enhance unsupervised domain adaptation with contrastive learning and with weak supervision when label proportions are available.
\end{abstract}



\maketitle

\section{Introduction}

Humans hunger to recognize and understand behavior patterns. An individual's activities affect their family, the workplace, society, and the environment. Human behavior is also a primary influence of a person's own health and wellness \cite{marteau2012,healthypeople}. Because behavior is reflected in mobile sensor data and such data are being amassed from commercial devices, researchers are increasingly tackling the problem of automating human activity recognition (HAR) from time series sensor data. However, existing algorithmic methods and datasets are typically focused on modeling repetitive, scripted movements that are performed in controlled settings \cite{dang2020,chen2022}. Now that these HAR systems have demonstrated success in constrained conditions, the next step is to design algorithmic methods and datasets for in-the-wild activity recognition.

Modeling and recognizing human activities in natural settings introduces numerous challenges not found in controlled settings. First, ground-truth labels are not as readily available. While the activity category is specified for scripted activities, labeling activities in the wild is burdensome, sometimes causes interruptions in the activities being modeled, and is frequently subject to error. Corresponding recognition techniques therefore need to handle cases where labels are not available for all individuals.

Second, modeling behavior that is performed in natural settings must reflect the tremendous variability that exists in human behavior. Many machine learning algorithms are designed based on the closed-world assumption, where the test data distribution is assumed to be similar to the training data distribution. In human behavior data, this assumption does not hold. In fact, variability is considered by some to be the most conspicuous characteristic of human behavior \cite{dodge1931}. The change in distribution may lead to a drop in the performance of activity recognition algorithms. Hence, activity recognition models need to capture or adapt to these individual differences. Such differences occur not only between individuals but within a single person's behavior at distinct time points. Behavioral facets fluctuate with changes in physiology, the external environment, and life course \cite{nielsen2018,zhao2018}. The nonstationarity of sensor-based behavior time series data further complicates the ability to model activities, particularly in the absence of regularly-timed ground-truth labels.

We hypothesize that the performance of mobile activity recognition (MAR) in natural settings can be boosted by unsupervised domain adaptation. Adversarial learning, which utilizes a two-player game to produce a domain-invariant feature representation, has become a commonplace tool in domain adaptation \cite{ganin2016jmlr}.
We further postulate that by augmenting adversarial learning with contrastive learning, which is a technique pulling similar examples together in the representation space and pushing dissimilar examples apart \cite{khosla2020supervisedcontrastive}, we can improve domain adaptation performance in the face of the increased adaptation difficulty due to the tremendous variability between people and nonstationarity in behavior over time. Finally, we hypothesize that in such naturalistic and highly variable settings, further gains may be achieved through leveraging additional weak supervision information in the form of label proportions (if this information is available), which may be easier to collect from participants than additional labeled data. For example, such proportions can be gathered through self-reported answers to questions such as ``How many hours did you sleep last night?'' \cite{wilson2020codats}. While we do not collect this data in our datasets, we measure the performance gains achievable by utilizing simulated self-reported data, computed from the labeled data we collect.

In this paper, we present an unscripted mobile human activity recognition dataset and measure the performance gains achievable by utilizing time series domain adaptation algorithms. To achieve our goal of obtaining a model that performs well in natural settings, we collect a variety of sensor data from smartwatches worn by participants, all of whom are occasionally prompted to label what activity they are performing throughout everyday life. To cover the wide variability between people, we collect data from two different age groups, each of which consists of data from a large number of participants: one dataset component with 15 younger adult participants and another dataset component with 30 older adult participants. Additionally, to enable addressing nonstationarity in behavior over time, we collect each of these datasets over a period of time: the younger adult data were collected over two weeks separated by a month, and the older adult data were collected over four consecutive weeks.

On each of these datasets, we evaluate two time series domain adaptation methods. First, we analyze domain adaptation results in the context of cross-person adaptation, where we train a model on data from one set of participants and adapt the model to work well on a different participant's data. Cross-person adaptation is the most commonly-studied case of adaptation for human activity recognition \cite{wilson2020codats,caldaarxiv}. Second, because in the creation of our datasets we collected data over a period of time, we are able to measure performance degradation of the models as the training and testing data are separated by greater lengths in time. This concept shift over time that occurs in these datasets collected in real-world settings, indicative of the nonstationarity of human behavioral data, motivates the need for cross-time domain adaptation. We repurpose time series domain adaptation methods for this cross-time domain adaptation problem and analyze the results. While domain adaptation methods yield an improvement over no adaptation, they still struggle to yield satisfactory performance on these more challenging datasets we present, motivating the need for future time series domain adaptation work.

\vspace{1.0ex}

In this paper, we describe and analyze the performance of two time series domain adaptation techniques for cross-person adaptation. We evaluate the methods on a new mobile activity recognition dataset\footnote{https://github.com/WSU-CASAS/smartwatch-data}. Data collection spans multiple weeks, thus we determine the extent of concept drift in the data and analyze the performance of cross-time adaptation as well. For both cross-person and cross-time adaptation problems, we simulate weak supervision in the form of label proportions and measure the time series domain adaptation performance gains achievable by utilizing this additional form of information.

\section{Related Work}

\subsection{Activity Recognition}

Modeling and recognizing activities from time series sensor data has become a popular research topic. Diverse classical machine learning methods have been explored, including decision tree, nearest neighbor,
support vector machine, random forest, clustering, and Bayesian classifiers \cite{demrozi2020}. However, such methods often rely on manual definition and extraction of discriminating features from the raw sensor readings. More recently, deep network architectures have been designed to model activities from the raw sensor data. Data are extracted from sensors that include triaxial accelerometers and gyroscopes, magnetometers, and heart rate. Deep learners have achieved strong recognition performance for movement-driven activity categories. Examples of these categories include body postures and repetitive movements such as standing, walking, sitting, running, lying, climbing, eating, writing, dribbling, and falling, as well as grinding, drilling, and sanding \cite{hassan2018}.

While most recent HAR research restricts activity categories to simple movements, some researchers have investigated approaches to modeling composite activities, representing combinations of simple movements. For example, Vepakomma et al. \cite{vepakomma2015} model combinations of movements and location types (e.g., walk indoor versus run outdoor) as well as semantically complex activities (e.g., clean, cook). They ease the learning task by introducing handcrafted features into the model. Alternatively, Peng et al. \cite{peng2018} use a method which is effective when complex activities can be defined as a sequence of simple movements. This approach defines a hierarchical structure to first model simple activities, then learn combinations of the basic categories.

To evaluate these existing mobile human activity recognition approaches, researchers have created and shared HAR datasets. These popular resources include
UCI HAR \cite{anguita2013public}, UCI HHAR \cite{stisen2015smartdevices}, WISDM AR \cite{kwapisz2011wisdmar}, and WISDM AT \cite{lockhart2011wisdmat}. In keeping with common methodologies, these datasets typically contain sensor data for activities that are scripted, performed uniformly by multiple participants, and, with the exception of WISDM AT, collected in a laboratory setting. We are interested in further extending the diversity of modeled activities. These additional categories reflect activities found in common questionnaires such as the American Time Use Survey \cite{ATUS} (e.g., eat, sleep, work, relax, household chores, exercise). We also include categories found in the list of basic and instrumental activities of daily living \cite{guo2020} because of their relevance for clinical applications (e.g., hygiene, travel, cook, errands, chores).  These activity categories may be more relevant for downstream tasks such as behavior pattern and trend analysis, health monitoring and intervention, and activity-aware service provisioning.

\subsection{Domain Adaptation}

Numerous domain adaptation techniques have been developed \cite{wilson2020survey}. However, only limited progress has been made toward designing robust domain adaptation approaches for time series data. Among these existing approaches are R-DANN and VRADA \cite{purushotham2017variational}, variational domain-adversarial methods that utilize a variational recurrent neural network (RNN) as the feature extractor. While the RNN did yield promising adaptation results, other prior work \cite{wilson2020codats} indicated that a 1D convolutional neural network outperformed RNNs on a variety of time-series adaptation problems. Another recent approach, AdvSKM \cite{liu2021advskm}, minimizes a maximum mean discrepancy metric embedded in a spectral kernel network. ContrasGAN \cite{sanabria2021contrasgan} also attacks the HAR domain adaptation problem. In this case, a bi-directional generator adversarial network performs heterogeneous feature transfer and contrastive learning. Among these, we select CoDATS \cite{wilson2020codats} as a baseline method for comparison because of its ability to perform robust unsupervised domain adaptation and incorporate weak supervision.

\section{CASAS MAR Dataset}

To evaluate our algorithms and support further development of mobile activity recognition,
we create a new mobile activity recognition dataset, called CASAS MAR. This dataset consists of smartwatch data collected while participants perform their non-scripted daily routine in real-world settings. In contrast to the publicly-available human activity recognition datasets that are used in prior time series domain adaptation work  \cite{anguita2013public,stisen2015smartdevices,kwapisz2011wisdmar,lockhart2011wisdmat}, this new dataset reflects the variability of movements and activity strategies that are found in natural settings.

We designed a watchOS app to collect sensor data from the Apple watch accelerometer, gyroscope, and location services. Participants are queried at periodic intervals to identify their current activity. The dataset contains two components. CASAS MAR/YA represents data collected from younger adults and CASAS MAR/OA represents data collected from older adults. These two components allow us to examine differences in labeling strategies and activity recognition performance between the two groups.

\subsection{CASAS MAR/YA}
In the first dataset component, data were collected from 15 younger adult (college student) participants (mean age 21.0, standard deviation=2.2) over two separate weeks. The second week of data collection occurred one month after the first week started. The Apple Watch (Series 2) app continuously collected accelerometer and gyroscope data at 50 Hz when not on the charger and collected location data every 5 minutes. Every 30 minutes, the app would prompt the participants to label their current activity from among six categories: $cook$, $eat$, $hygiene$, $work$, $exercise$, and $travel$. Participants had a choice to ignore the prompt when performing an activity such as driving that inhibited responding to the prompt.

In contrast to the UCI HHAR dataset, where Stisen et al. \cite{stisen2015smartdevices} only sample either accelerometer or gyroscope sensors, our dataset includes accelerometer, gyroscope, and location readings. This expanded sensor set enables us to analyze values for 3-axis user acceleration, 3-axis rotation rate, yaw, pitch, roll, latitude, longitude, altitude, speed, and the corresponding timestamp. We stored the sensor data on the watches using Google's Protocol Buffers (Protobuf) \cite{popic2016performance}, which enables defining both an efficient storage format (each watch has space for over 2 weeks of data in this format) and provides extensibility and backwards compatibility.

The sampling frequency of 50 Hz was chosen because past work used this frequency for activity recognition \cite{anguita2013public,shoaib2015towards,shoaib2016complex,shoaib2015survey,cope2009estimating}. After data collection, we downsampled to 10 Hz for consistency between dataset components while maintaining classification performance \cite{shoaib2015towards}. In the case of location data, we sampled once every 5 minutes. This represented the maximum sampling frequency that ensured the app could continuously collect data for up to 18 hours on a single charge. This constraint allowed participants to wear the device from when they woke up to when they went to bed each day.

The domain adaptation networks are not designed to fuse time series data arriving at non-uniform rates. To provide uniform data, we maintain the last-recorded location values between readings. When a participant provides an activity label at time $t$, the label is applied to all readings that were collected from time $t$ minus 5 minutes to time $t$. We then segment the time series into non-overlapping windows of 128 time steps. This size is selected to be consistent with other publicly-available datasets. We do not encode raw latitude/longitude/altitude location values because 1) they do not generalize effectively to new subjects and 2) they introduce a privacy risk. Instead, we perform a reverse geocode lookup to identify the location type that is associated with the location coordinate using the OpenStreetMap Nominatim database \cite{nominatim}. The values for location type are other, house, road, service, work, and attraction. We one-hot encode the resulting location categories for our feature vector. Since activities of daily living may be affected by time and date, we additionally include the day of week, hour of the day, and minutes past midnight as features.

\subsection{CASAS MAR/OA}

In the second dataset component, data were collected from 30 older adult participants (mean age 70.7, standard deviation=7.8) for one month.
To accommodate the preferences of this group, each participant was prompted to label their activity at most 10 times per day. Prompts were given during waking hours (e.g., between the hours of 9am and 9pm). The reduced query frequency lightened the burden of the study on participants and ensured participant responsiveness for the entire month. The prompt hours are selected to increase instance diversity over the waking hours, thus a prompt is timed for an hour where the number of labeled instances is $\leq$ the median number of labels across all waking hours.

In contrast with the first dataset component, we collected data using Apple Watch (Series 5), which provides sufficient storage space for the increased study duration. We additionally collected data directly at 10 Hz rather than downsampling after data collection. Finally, we also expanded the label set for this study to include a total of nine activity categories: $errands$, $exercise$, $hobby$, $hygiene$, $mealtime$, $relax$, $sleep$, $travel$, and $work$.

\section{Methodology}

Our long-term goal is to perform robust activity recognition from mobile sensors, even amidst changes in the data distribution. We study two techniques to this end.

First, we study the problem of unsupervised domain adaptation. In domain adaptation, we define two domains: one or more source domains and a target domain. For unsupervised domain adaptation, the source domains consist of labeled data, and the target domain only requires unlabeled data. For example, we may collect labeled data from a number of ``source'' participants 1-5 with the goal of producing a model that performs well for ``target'' person 6. We can reduce the burden of creating this model for person 6 by not requiring this target participant to label data, but instead only wear the watch for a period of time collecting unlabeled data. We can measure the performance of this technique by selecting some portion of a given MAR dataset to be ``sources'' and another portion to be ``target'', e.g., for cross-person adaptation, selecting based on participant identifier, or, for cross-time adaptation, selecting based on timestamp.

Second, we enhance domain adaptation with weak supervision \cite{wilson2020codats}, a technique that leverages additional information about the target domain, if available. Leveraging such weak supervision is an alternative method of obtaining improved domain adaptation performance in a manner that may reduce the data collection burden on participants. Rather than labeling data instances, the target participant instead reports what proportion of the time they were performing different activities, e.g., how many hours a day they exercised or slept. While we did not collect this data in our study, we can simulate this technique by measuring the label proportions on the target domain (while taking care not to make use of the labels directly, which would violate the assumptions of unsupervised domain adaptation).

\subsection{Problem Setup}

Following the formulation in CoDATS \cite{wilson2020codats}, we define several distributions in terms of the input data $X$ and label space $L$ consisting of labels $Y = \{ 1, 2, \dots, L \}$: $n$ source domain distributions $\mathcal{D}_{S_i}$ and a target domain distribution $\mathcal{D}_T$. During training, we draw samples i.i.d. from each labeled source domain distribution and also i.i.d. from the unlabeled marginal target distribution of $\mathcal{D}_T$ over $X$, yielding $s_i$ source samples and $t$ target samples $T$:
\begin{equation}\label{eq:msda1}
    S_i = \{ (\textbf{x}_j, y_j) \}^{s_i}_{j=1} \sim \mathcal{D}_{S_i}, \quad \forall i \in \{ 1, 2, \dots, n \}
\end{equation}
\begin{equation}\label{eq:msda2}
    T = \{ (\textbf{x}_j) \}^t_{j=1} \sim \mathcal{D}_T^X
\end{equation}

During testing, we evaluate the model on samples drawn i.i.d. from the target distribution $\mathcal{D}_T$, yielding $t$ target samples not seen during training:
\begin{equation}\label{eq:msda3}
  T_{test} = \{ (\textbf{x}_j, y_j) \}^t_{j=1} \sim \mathcal{D}_T
\end{equation}

For the problem of domain adaptation with weak supervision, we are additionally provided with the target-domain label proportions, a discrete distribution:
\begin{equation}\label{eq:labelproportions}
    Y_{true}(y) = P(Y=y) = p_y, \quad \forall y \in \{ 1, 2, \dots, L \}
\end{equation}

Because we focus here on time series mobile activity recognition data, each $X$ represents a multivariate time series. A univariate time series can be represented by a list of $H$ ordered real values $X = [x_1, x_2, \dots, x_H]$. Then, a multivariate time series can be represented by a list of $K$ univariate time series $X = [X^1, X^2, \dots, X^K]$. Our datasets consist of sensor data, e.g., given accelerometer $x$, $y$, and $z$ values, we obtain a series of these readings for each univariate axis.

\subsection{Time Series Domain Adaptation Methods}

\begin{figure}
  \centering
  \includegraphics[width=1.0\linewidth]{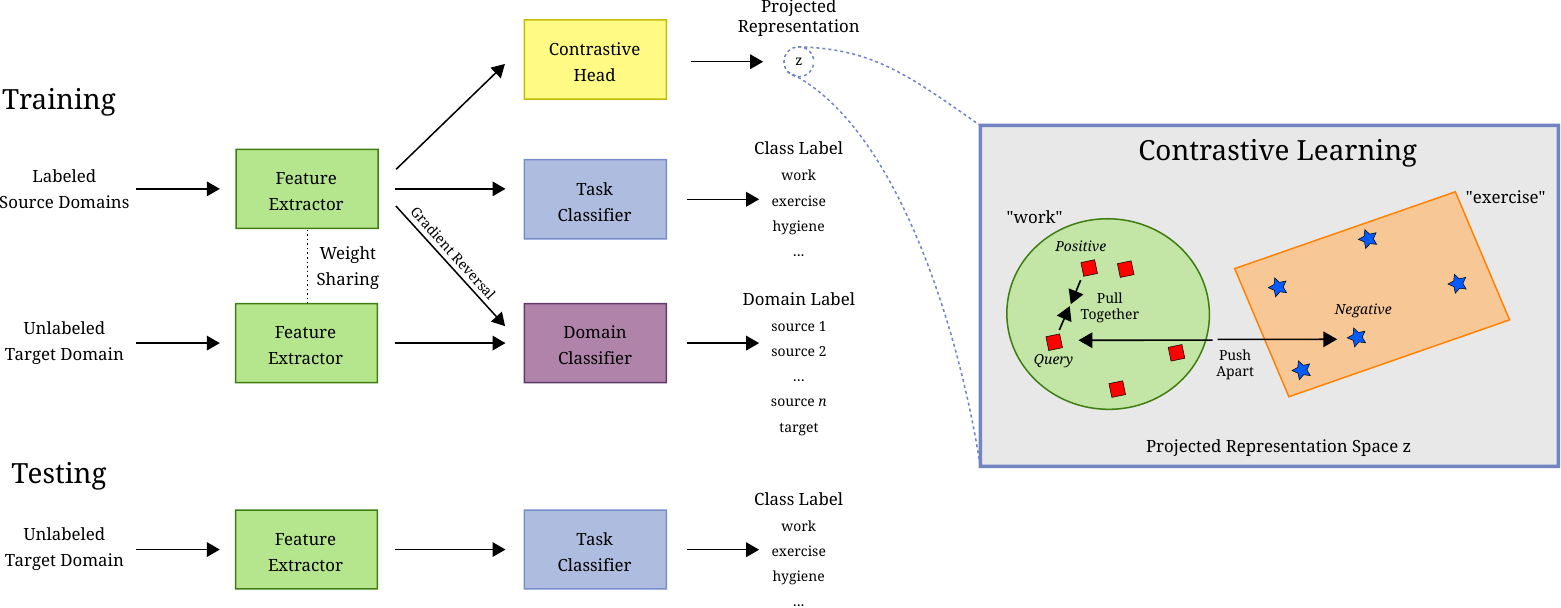}
  \caption{Time Series Domain Adaptation Methods: (1) CoDATS consists of the feature extractor, task classifier, and domain classifier. (2) CALDA additionally adds a contrastive head to which a supervised contrastive loss is applied. This contrastive strategy pulls together same-labeled examples and pushes apart different-labeled examples.}
  \label{fig:methods}
\end{figure}

We select two time series domain adaptation methods to boost mobile activity recognition performance. The first is CoDATS \cite{wilson2020codats}, and the second is called CALDA. Both methods perform unsupervised domain adaptation and can incorporate weak supervision. These two methods illustrated in Figure~\ref{fig:methods}. CoDATS uses a domain adversarial neural network \cite{ganin2016jmlr}. Building on the CoDATS architecture, CALDA additionally incorporates contrastive learning \cite{khosla2020supervisedcontrastive}. Because both of these methods support leveraging weak supervision information, such as target domain label proportions that participants may be able to self-report easier than labeling additional data instances, we include the two weak supervision variations of these methods, CoDATS-WS and CALDA-WS, as well.

\subsubsection{CoDATS}

The CoDATS time series domain adaptation method consists of a time series compatible model architecture with three components: a feature extractor, a domain classifier, and a task classifier \cite{wilson2020codats}. The feature extractor is based on 1D convolutions, converting the raw time series data into a learned representation space. The resulting feature representation is then input into the domain and task classifiers. The multi-class domain classifier acts as an adversary, learning to classify which domain the data originated from. By negating this gradient, CoDATS can learn a domain-invariant representation that enables improved transfer from one or more source domains to the target domain.

Formally, given a feature extractor $F$ with parameters $\theta_f$, a domain classifier $D$ with parameters $\theta_d$, a task classifier $C$ with parameters $\theta_c$, a gradient reversal layer $\mathcal{R}$ \cite{ganin2016jmlr}, cross-entropy losses $\mathcal{L}_d$ for $D$ and $\mathcal{L}_y$ for $C$, and domain labels $d_{S_i}$ for the source domains and $d_T$ for the target domain, we obtain the following CoDATS training objective \cite{wilson2020codats}:
\begin{equation}\label{eq:grlFormulation}
\begin{split}
\argmin_{\theta_f,\theta_c,\theta_d} \quad \sum_{i=1}^{n} \Big[ \mathbb{E}_{(\mathbf{x},y) \sim \mathcal{D}_{S_i}} \big[ & \mathcal{L}_y\left(C(F(\mathbf{x})), y\right) + \\
& \mathcal{L}_d\left(D(\mathcal{R}(F(\mathbf{x}))), d_{S_i}\right) \big] \Big] + \\
& \mathbb{E}_{\mathbf{x} \sim \mathcal{D}_T^X} \big[ \mathcal{L}_d\left(D(\mathcal{R}(F(\mathbf{x}))), d_T\right) \big]
\end{split}
\end{equation}

\subsubsection{CALDA}

Building upon CoDATS, CALDA additionally incorporates contrastive learning \cite{caldaarxiv}. This domain adaptation algorithm adds an additional contrastive head component to the network architecture, to which an InfoNCE supervised contrastive loss is applied \cite{oord2018infonce,khosla2020supervisedcontrastive}. In the representation space of this new contrastive head, the supervised contrastive loss pulls together source domain examples that have the same label and pushes apart source domain examples that have different labels. This process is illustrated in Figure~\ref{fig:methods} (right). Such feature-space modification is performed via pairs of examples in the training data. First, a query example is selected. Then, a positive set is constructed consisting of examples with the same label as the query, and a negative set is constructed consisting of examples with different labels than the query. Finally, the contrastive loss in Equation~\ref{eq:contrastive} is minimized for each of the query examples, which pulls the positive set of examples closer to the query and pushes the negative set of examples away. Through this process, CALDA produces a feature representation that generalizes better among the multiple domains, and thereby improves target domain performance.

Formally, given a projected representation $z$ of the query, the positive set $P$, the negative set $N$, a temperature $\tau$, and cosine similarity $\text{sim}(z_1, z_2) = \frac{z_1^T z_2}{\|z_1\|\|z_2\|}$, we can define an InfoNCE supervised contrastive loss \cite{oord2018infonce,khosla2020supervisedcontrastive}:
\begin{equation}\label{eq:contrastive}
    \mathcal{L}_c(z, P, N) = \frac{1}{|P|} \sum_{z_p \in P} \left[ - \log \left( \frac{\exp\left({\frac{\text{sim}\left(z, z_p\right)}{\tau}}\right)}{\sum_{z_k \in N \cup \{z_p\}} \exp\left({\frac{\text{sim}\left(z, z_k\right)}{\tau}}\right)} \right) \right]
\end{equation}

Next, we define the query, positive, and negative sets using set-builder notation. Given an auxiliary set $K$ containing the input $x$, class label $y$, and domain label $d$, the set of labeled examples $S_i \sim \mathcal{D}_{S_i}$ from each of the $n$ source domains, and the projected representation from the contrastive head $Z$ with parameters $\theta_z$, we define the query set $Q_{S_i}$ for each source domain, positive set $P_{d_q, y_q}$ for a query with domain $d_q$ and label $y_q$, and negative set $N_{d_q, y_q}$ similarly. Note that the difference between the positive and negative sets is whether the positive/negative label matches the query's label or not.
\begin{equation}
  K = \{ (x, y, d) \mid (x, y) \in S_i, i \in \{ 1, 2, \dots, n \}, d = d_{S_i} \}
\end{equation}
\begin{equation}
  Q_{S_i} = \{ (z, y) \mid (x, y, d) \in K, z = Z(F(x)), d = d_{S_i} \}
\end{equation}
\begin{equation}
  P_{d_q, y_q} = \{ z \mid (x, y, d) \in K, z = Z(F(x)), y = y_q \}
\end{equation}
\begin{equation}
  N_{d_q, y_q} = \{ z \mid (x, y, d) \in K, z = Z(F(x)), y \neq y_q \}
\end{equation}

Then, given a feature extractor $F$ with parameters $\theta_f$, a contrastive head $Z$ with parameters $\theta_z$, the query set from each source domain $Q_{S_i}$, and positive and negative sets $P_{d,y}$ and $N_{d,y}$ for each query from domain $d$ and with label $y$, we obtain the contrastive training objective:
\begin{equation}\label{eq:contrastive_objective}
    \argmin_{\theta_f,\theta_z} \sum_{i=1}^n \left[ \frac{1}{|Q_{S_i}|} \sum_{(z_q, y_q) \in Q_{S_i}} \mathcal{L}_c(z_q, P_{d_{S_i}, y_q}, N_{d_{S_i}, y_q}) \right]
\end{equation}

\subsubsection{Weak Supervision}

Both CoDATS and CALDA support leveraging weak supervision information if available. By incorporating weak supervision methods into domain adaptation, we can determine the level of performance improvement from utilizing the weak supervision information. To evaluate the benefit of weak supervision, in CoDATS-WS and CALDA-WS we add a KL divergence regularization term to the training objective that aligns the predicted task classifier's label distribution for the target domain's unlabeled data with the known target domain label proportion distribution \cite{wilson2020codats}.

Formally, given a feature extractor $F$ with parameters $\theta_f$, a task classifier $C$ with parameters $\theta_c$, and the known target-domain label proportion distribution $Y_{true}$, we add the following KL divergence regularization term to the training objective:
\begin{equation}\label{eq:ws}
  \argmin_{\theta_f,\theta_c} \left[ \kld*{Y_{true}}{\mathbb{E}_{x \sim \mathcal{D}_T^X} \big[ C(F(x)) \big]} \right]
\end{equation}

\begin{table}
  \centering
  \caption{Per-Class Variation in HAR Datasets for Average Euclidean Distance (ED) and Average KL Divergence (KL)}
  \label{table:variation}
  \begin{scriptsize}
  {\renewcommand{\arraystretch}{1.4}
  \begin{tabular}{ccccccc}
\toprule
Dataset & Dim & Domains & Examples & Class & ED & KL \\
\midrule
MAR/ & 20 & 15 & 2122 & Cook & 66.6 $\pm$ 23.8 & 4551 $\pm$ 2333 \\
YA & 20 & 15 & 2608 & Eat & 64.1 $\pm$ 28.5 & 4301 $\pm$ 2276 \\
& 20 & 15 & 1798 & \textit{Exercise} & 83.9 $\pm$ 28.3 & 5165 $\pm$ 2444 \\
& 20 & 15 & 1529 & \textit{Hygiene} & 70.5 $\pm$ 25.5 & 4722 $\pm$ 2399 \\
& 20 & 15 & 2802 & \textit{Travel} & 89.0 $\pm$ 31.7 & 5114 $\pm$ 2206 \\
& 20 & 15 & 13955 & \textit{Work} & 49.2 $\pm$ 18.4 & 3279 $\pm$ 2117 \\
& & & & {\bf Average} & {\bf 70.6 $\pm$ 26.4} & {\bf 4522 $\pm$ 2299} \\ 
\hline
MAR/ & 20 & 25 & 7153 & Errands & 83.1 $\pm$ 29.4 & 4651 $\pm$ 2152 \\
OA & 20 & 25 & 5707 & \textit{Exercise} & 81.0 $\pm$ 25.9 & 4684 $\pm$ 2300 \\
& 20 & 25 & 7005 & Hobby & 58.0 $\pm$ 23.0 & 3833 $\pm$ 2185 \\
& 20 & 25 & 5222 & \textit{Hygiene} & 64.2 $\pm$ 20.6 & 3792 $\pm$ 2110 \\
& 20 & 25 & 13723 & Mealtime & 58.3 $\pm$ 17.6 & 3786 $\pm$ 2167 \\
& 20 & 25 & 14474 & Relax & 51.9 $\pm$ 18.2 & 3426 $\pm$ 2244 \\
& 20 & 25 & 1589 & Sleep & 47.6 $\pm$ 18.0 & 3126 $\pm$ 2327 \\
& 20 & 25 & 4940 & \textit{Travel} & 66.4 $\pm$ 26.2 & 4104 $\pm$ 2038 \\
& 20 & 25 & 10382 & \textit{Work} & 56.0 $\pm$ 16.8 & 4016 $\pm$ 2139 \\
& & & & {\bf Average} & {\bf 62.9 $\pm$ 22.2} & {\bf 3935 $\pm$ 2186} \\ 
\hline
UCI & 9 & 30 & 1237 & Laying & 7.4 $\pm$ 5.4 & 350 $\pm$ 592 \\
HAR & 9 & 30 & 1133 & Sitting & 4.4 $\pm$ 2.5 & 227 $\pm$ 350 \\
& 9 & 30 & 1213 & Standing & 3.2 $\pm$ 1.5 & 107 $\pm$ 134 \\
& 9 & 30 & 1100 & Walking & 14.8 $\pm$ 2.7 & 798 $\pm$ 220 \\
& 9 & 30 & 899 & Downstairs & 18.3 $\pm$ 3.1 & 965 $\pm$ 273 \\
& 9 & 30 & 984 & Upstairs & 16.5 $\pm$ 3.0 & 899 $\pm$ 307 \\
& & & & {\bf Average} & {\bf 10.8 $\pm$ 3.3} & {\bf 558 $\pm$ 344} \\ 
\hline
UCI & 3 & 9 & 10068 & Bike & 21.1 $\pm$ 5.6 & 805 $\pm$ 545 \\
HHAR & 3 & 9 & 10854 & Sit & 14.8 $\pm$ 8.2 & 455 $\pm$ 596 \\
& 3 & 9 & 8808 & Stairsdown & 32.8 $\pm$ 7.0 & 840 $\pm$ 420 \\
& 3 & 9 & 9713 & Stairsup & 29.3 $\pm$ 6.7 & 768 $\pm$ 442 \\
& 3 & 9 & 10093 & Stand & 8.9 $\pm$ 5.8 & 381 $\pm$ 609 \\
& 3 & 9 & 11953 & Walk & 29.6 $\pm$ 5.0 & 863 $\pm$ 391 \\
& & & & {\bf Average} & {\bf 22.8 $\pm$ 6.5} & {\bf 685 $\pm$ 508} \\ 
\hline
WISDM & 3 & 33 & 492 & Downstairs & 21.3 $\pm$ 4.5 & 815 $\pm$ 358 \\
AR & 3 & 33 & 1665 & Jogging & 35.5 $\pm$ 3.8 & 1078 $\pm$ 361 \\
& 3 & 33 & 293 & Sitting & 16.0 $\pm$ 8.2 & 548 $\pm$ 819 \\
& 3 & 33 & 239 & Standing & 10.9 $\pm$ 5.6 & 393 $\pm$ 605 \\
& 3 & 33 & 605 & Upstairs & 20.8 $\pm$ 4.7 & 768 $\pm$ 369 \\
& 3 & 33 & 1972 & Walking & 22.6 $\pm$ 3.3 & 866 $\pm$ 295 \\
& & & & {\bf Average} & {\bf 21.2 $\pm$ 5.3} & {\bf 745 $\pm$ 503} \\ 
\hline
WISDM & 3 & 51 & 2029 & Jogging & 32.8 $\pm$ 7.5 & 1023 $\pm$ 554 \\
AT & 3 & 51 & 1194 & Lying Down & 12.9 $\pm$ 12.2 & 374 $\pm$ 815 \\
& 3 & 51 & 2718 & Sitting & 20.3 $\pm$ 15.9 & 788 $\pm$ 986 \\
& 3 & 51 & 173 & Stairs & 28.4 $\pm$ 9.9 & 849 $\pm$ 549 \\
& 3 & 51 & 1116 & Standing & 19.7 $\pm$ 15.9 & 551 $\pm$ 697 \\
& 3 & 51 & 3330 & Walking & 26.3 $\pm$ 7.4 & 919 $\pm$ 517 \\
& & & & {\bf Average} & {\bf 23.4 $\pm$ 12.0} & {\bf 751 $\pm$ 707} \\ 
\bottomrule
\end{tabular}

  }
  \end{scriptsize}
\end{table}

\section{Experimental Results}

We experimentally evaluate the impact of unsupervised domain adaptation on the CASAS MAR mobile activity recognition problem. The CASAS MAR dataset provides multiple opportunities for domain adaptation. First, we collect data from multiple participants. The methods and movements each person uses to perform activities is highly variable, so we can view each person's data as a separate domain for adaptation. If successful, the domain adaptation will be effective at recognizing activities for new persons who cannot provide ground-truth activity labels. Second, we view data collected at distinct time points as separate domains. Because behavior changes over time, a model that is trained from ground-truth data at one timepoint may not generalize well enough to adequately recognize the same activity classes performed at a different time. Thus, we can employ unsupervised domain adaptation to adapt models to new seasons, years, or other timeframes.

In these experiments, we evaluate the CoDATS and CALDA methods, with and without weak supervision.
Following the CoDATS training methodology, for the cross-person experiments, we select 3 random sets of source domains for each of 10 different target domains for each value of $n$ (the number of source domains). The error bars represent the average of the standard deviation across the set of 3 random sets of source domains. For the cross-time experiments, we perform adaptation from one week to another week, averaged over each of the different participants. The error bars represent standard deviation.

We additionally include a ``No Adaptation'' lower bound baseline, which enables us to measure the performance improvement achieved by utilizing domain adaptation. Finally, we include an upper bound ``Train on Target''. This model is trained using the labels from the training set of the target domain. While this method ``cheats'' by accessing labeled target training data, it provides an upper bound estimate of how well a model could perform if ample target domain labels were available. We report performance in each based based on AUC values for instances in the target domain.

\subsection{Variability of Human Behavior}

This work is motivated by the need to perform activity recognition in naturalistic, unscripted settings. We expect that data collected while participants perform normal daily activities will exhibit much greater variation than is the case for scripted activities. To experimentally compare these datasets,
in Table~\ref{table:variation}, we measure the variation among examples of each class for the CASAS MAR dataset as well as other publicly-available HAR datasets. We compute the distances between the raw data of all training set examples in each class using two measures: Euclidean distance and KL divergence. We observe differences between labels within CASAS MAR/YA and CASAS MAR/OA and also in the shared labels between these two dataset components (denoted by italics). The differences are indicative of differences in behavior between these groups. We note that the observed distances and standard deviations are much larger in CASAS MAR than in the prior human activity recognition datasets shown in the table, despite all of these datasets being processed in a similar manner. The increased variability is an indication that MAR will be challenging and may benefit from domain adaptation. Given the more challenging nature of CASAS MAR due to the increased variability in behavior, we next proceed to measure the benefit of utilizing time series domain adaptation methods to improve model performance under the individual differences and temporal differences present in natural activity recognition data.

\begin{table}
  \centering
  \caption{Cross-Person Younger Adults (AUC)}
  \label{table:person_younger}
  \begin{scriptsize}
  {\renewcommand{\arraystretch}{1.4}
  \begin{tabular}{cccccc}
\toprule
$n$ & No Adaptation & CoDATS & CoDATS-WS & \textit{CALDA} & \textit{CALDA-WS} \\
\midrule
2 & 0.686 $\pm$ 0.039 & 0.707 $\pm$ 0.035 & 0.722 $\pm$ 0.028 & 0.750 $\pm$ 0.030 & \textbf{0.777 $\pm$ 0.026} \\
5 & 0.707 $\pm$ 0.033 & 0.742 $\pm$ 0.025 & 0.749 $\pm$ 0.019 & 0.790 $\pm$ 0.015 & \textbf{0.818 $\pm$ 0.012} \\
8 & 0.717 $\pm$ 0.028 & 0.759 $\pm$ 0.018 & 0.763 $\pm$ 0.015 & 0.801 $\pm$ 0.017 & \textbf{0.828 $\pm$ 0.013} \\
11 & 0.724 $\pm$ 0.017 & 0.779 $\pm$ 0.013 & 0.777 $\pm$ 0.012 & 0.811 $\pm$ 0.009 & \textbf{0.837 $\pm$ 0.008} \\
14 & 0.735 $\pm$ 0.000 & 0.790 $\pm$ 0.000 & 0.775 $\pm$ 0.000 & 0.813 $\pm$ 0.000 & \textbf{0.835 $\pm$ 0.000} \\
\hline
Avg & 0.714 $\pm$ 0.024 & 0.755 $\pm$ 0.018 & 0.757 $\pm$ 0.015 & 0.793 $\pm$ 0.014 & \textbf{0.819 $\pm$ 0.012} \\
\bottomrule
\end{tabular}

  }
  \end{scriptsize}
\end{table}

\begin{table}
  \centering
  \caption{Cross-Person Older Adults (AUC)}
  \label{table:person_older}
  \begin{scriptsize}
  {\renewcommand{\arraystretch}{1.4}
  \begin{tabular}{cccccc}
\toprule
$n$ & No Adaptation & CoDATS & CoDATS-WS & \textit{CALDA} & \textit{CALDA-WS}  \\
\midrule
2 & 0.637 $\pm$ 0.043 & 0.585 $\pm$ 0.027 & 0.617 $\pm$ 0.021 & 0.700 $\pm$ 0.043 & \textbf{0.777 $\pm$ 0.019} \\
7 & 0.637 $\pm$ 0.035 & 0.604 $\pm$ 0.022 & 0.625 $\pm$ 0.018 & 0.716 $\pm$ 0.031 & \textbf{0.787 $\pm$ 0.022} \\
12 & 0.650 $\pm$ 0.037 & 0.614 $\pm$ 0.025 & 0.646 $\pm$ 0.026 & 0.733 $\pm$ 0.027 & \textbf{0.796 $\pm$ 0.020} \\
17 & 0.663 $\pm$ 0.024 & 0.624 $\pm$ 0.016 & 0.655 $\pm$ 0.020 & 0.742 $\pm$ 0.020 & \textbf{0.806 $\pm$ 0.011} \\
22 & 0.670 $\pm$ 0.016 & 0.648 $\pm$ 0.019 & 0.677 $\pm$ 0.022 & 0.738 $\pm$ 0.010 & \textbf{0.806 $\pm$ 0.009} \\
\hline
Avg & 0.651 $\pm$ 0.031 & 0.615 $\pm$ 0.021 & 0.644 $\pm$ 0.022 & 0.726 $\pm$ 0.026 & \textbf{0.794 $\pm$ 0.016} \\
\bottomrule
\end{tabular}

  }
  \end{scriptsize}
\end{table}

\begin{table}
  \centering
  \caption{Cross-Time Younger Adults (AUC)}
  \label{table:time_younger}
  \begin{scriptsize}
  {\renewcommand{\arraystretch}{1.4}
  \begin{tabular}{cccccc}
\toprule
No Adaptation & CoDATS & CoDATS-WS & \textit{CALDA} & \textit{CALDA-WS} & Train/Target \\
\midrule
week 0 $\rightarrow$ 4 & & & & & \\
0.819 $\pm$ 0.034 & 0.810 $\pm$ 0.040 & 0.834 $\pm$ 0.053 & \textbf{0.872 $\pm$ 0.042} & 0.870 $\pm$ 0.055 & 0.996 $\pm$ 0.002 \\
week 4 $\rightarrow$ 0 & & & & & \\
0.792 $\pm$ 0.044 & 0.780 $\pm$ 0.055 & 0.791 $\pm$ 0.052 & \textbf{0.838 $\pm$ 0.044} & 0.832 $\pm$ 0.042 & 0.992 $\pm$ 0.004 \\
\hline
Avg & & & & & \\
0.806 $\pm$ 0.039 & 0.795 $\pm$ 0.047 & 0.812 $\pm$ 0.053 & \textbf{0.855 $\pm$ 0.043} & 0.851 $\pm$ 0.048 & 0.994 $\pm$ 0.003 \\
\bottomrule
\end{tabular}

  }
  \end{scriptsize}
\end{table}

\begin{table}
  \centering
  \caption{Cross-Time Older Adults (AUC)}
  \label{table:time_older}
  \begin{scriptsize}
  {\renewcommand{\arraystretch}{1.4}
  \begin{tabular}{cccccc}
\toprule
No Adaptation & CoDATS & CoDATS-WS & \textit{CALDA} & \textit{CALDA-WS} & Train/Target \\
\midrule
week 0 $\rightarrow$ 1 & & & & & \\
0.636 $\pm$ 0.073 & 0.600 $\pm$ 0.070 & 0.656 $\pm$ 0.066 & 0.706 $\pm$ 0.072 & \textbf{0.761 $\pm$ 0.076} & 0.988 $\pm$ 0.009 \\
week 0 $\rightarrow$ 2 & & & & & \\
0.630 $\pm$ 0.081 & 0.604 $\pm$ 0.073 & 0.643 $\pm$ 0.087 & 0.701 $\pm$ 0.084 & \textbf{0.747 $\pm$ 0.090} & 0.987 $\pm$ 0.007 \\
week 0 $\rightarrow$ 3 & & & & & \\
0.622 $\pm$ 0.108 & 0.591 $\pm$ 0.090 & 0.648 $\pm$ 0.089 & 0.670 $\pm$ 0.127 & \textbf{0.729 $\pm$ 0.096} & 0.990 $\pm$ 0.007 \\
week 1 $\rightarrow$ 0 & & & & & \\
0.633 $\pm$ 0.090 & 0.610 $\pm$ 0.067 & 0.661 $\pm$ 0.082 & 0.673 $\pm$ 0.088 & \textbf{0.736 $\pm$ 0.084} & 0.984 $\pm$ 0.010 \\
week 2 $\rightarrow$ 0 & & & & & \\
0.630 $\pm$ 0.098 & 0.608 $\pm$ 0.084 & 0.655 $\pm$ 0.108 & 0.663 $\pm$ 0.098 & \textbf{0.723 $\pm$ 0.089} & 0.984 $\pm$ 0.010 \\
week 3 $\rightarrow$ 0 & & & & & \\
0.615 $\pm$ 0.098 & 0.601 $\pm$ 0.078 & 0.630 $\pm$ 0.087 & 0.659 $\pm$ 0.096 & \textbf{0.717 $\pm$ 0.087} & 0.984 $\pm$ 0.010 \\
\hline
Avg & & & & & \\
0.627 $\pm$ 0.091 & 0.602 $\pm$ 0.077 & 0.649 $\pm$ 0.086 & 0.679 $\pm$ 0.094 & \textbf{0.735 $\pm$ 0.087} & 0.986 $\pm$ 0.009 \\
\bottomrule
\end{tabular}

  }
  \end{scriptsize}
\end{table}

\subsection{Cross-Person Adaptation}

The cross-person adaptation results are shown in Tables~\ref{table:person_younger} and \ref{table:person_older}. Training on the target yields an AUC value of 0.992 $\pm$ 0.001 for younger adults and 0.979 $\pm$ 0.003 for older adults. With younger adults, we find the domain adaptation approaches improve over No Adaptation in all cases. These results agree with prior cross-person time series domain adaptation work studying human activity recognition \cite{wilson2020codats,caldaarxiv}. With older adults, we find that CALDA always improves over No Adaptation, though CoDATS performs slightly worse. The decreased performance for CoDATS indicates the limitations of this baseline method when applied to increasingly challenging datasets. However, the successful adaptation by CALDA points to the benefit of leveraging contrastive learning in MAR domain adaptation.

\subsection{Cross-Time Adaptation}

The cross-time adaptation results are shown in Tables~\ref{table:time_younger} and \ref{table:time_older}. We observe a benefit from domain adaptation with CoDATS-WS, CALDA, and CALDA-WS for both younger and older adults. Due to the limited data available for the younger adults, we cannot perform adaptation over gradually increasing time gaps. However, in the single cross-time adaptation problem from the first week of data to the week a month later shown in Table~\ref{table:time_younger}, we do observe concept drift as illustrated by the No Adaptation performance. For the older adults, we collected data throughout an entire month, enabling measuring the concept drift with increasing time gap. As shown in Table~\ref{table:time_older}, performance in almost all cases degrades with increasing time gaps (with the exception of CoDATS and CoDATS-WS in the forward direction). This decreasing performance over time demonstrates the concept drift over time.

These results motivate the need for cross-time domain adaptation. Collected labeled data for a period of time (e.g., one week or one month) and then subsequently using that model for an extended period of time will result in sub-par performance. As shown in these results, with both younger adults and older adults, this performance is far below that of the upper bound Train on Target. However, some of this performance drop can be regained through performing unsupervised time series domain adaptation, using a combination of the original period of labeled data and additionally some unlabeled data from the more recent time period to produce an adapted model for the more recent time period.

\subsection{Younger vs. Older Adults}

In each of the cross-person and cross-time tables, if we compare the younger adult results with the older adult results, we observe that the older adult model performance is worse than the younger adult model performance. We believe there are two causes for this: increased variability in human behavior and decreased labeling density. In the Appendix, we validate that the difference in class labels is not the cause.

First, the younger adults were a more homogeneous group, largely consisting of students at one university. This means that there is limited variation for each activity label, e.g., each person's ``work'' label is likely related to homework and each person's ``travel'' likely consists of walking to classes. In contrast, the older adults may have vastly different ``work'' or ``travel'' labels depending on their occupation and where they live.

Second, we collected a higher density of labeled data for the younger adults by prompting each person to provide a label once every 30 minutes, due to the shorter duration of this study and the corresponding limited-time disruption to participant routines. On the other hand, with older adults, the study duration was over twice as long, so we limited the number of daily queries for activity labels.

\section{Conclusion}

While human activity recognition is a popular area of research, little effort has focused on recognizing complex behaviors in the wild. Creating robust recognition methods in naturalistic settings with limited ground truth labels benefits from leveraging unsupervised domain adaptation techniques. These methods are further improved by incorporating contrastive learning and weak supervision. To validate our methods and provide a resource for mobile activity recognition, we have collected two MAR dataset components consisting of data collected from both younger and older adults performing activities in natural settings. We demonstrated the increased variability present in these datasets when compared with existing publicly-available datasets, indicating the challenging nature of our datasets. Our experimental results, both for cross-person and cross-time adaptation, verify that MAR achieves performance gains by leveraging contrastive learning in addition to adversarial learning. Future work includes developing more sophisticated time series domain adaptation methods that can better address the large domain gaps present in such challenging realistic datasets.

\begin{acks}
The authors would like to thank Justin Frow and Maureen Schmitter-Edgecombe for their help with the data collection. Research reported in this publication was supported by the National Institutes of Health under award R25AG046114 and by the National Science Foundation under award 1954372.
\end{acks}

\bibliographystyle{ACM-Reference-Format}
\bibliography{references}

\appendix

\section{Additional Results}

\begin{table}
  \centering
  \caption{Cross-Person Younger Adults (Accuracy)}
  \label{table:person_younger_acc}
  \begin{scriptsize}
  {\renewcommand{\arraystretch}{1.4}
  \begin{tabular}{ccccccc}
\toprule
$n$ & No Adaptation & CoDATS & CoDATS-WS & \textit{CALDA} & \textit{CALDA-WS} & Train/Target \\
\midrule
2 & 42.1 $\pm$ 6.4 & 44.0 $\pm$ 6.1 & 46.9 $\pm$ 3.6 & 47.6 $\pm$ 5.1 & \textbf{50.8 $\pm$ 3.7} & 95.1 $\pm$ 0.3 \\
5 & 45.0 $\pm$ 5.3 & 46.8 $\pm$ 4.1 & 49.8 $\pm$ 2.5 & 52.3 $\pm$ 2.1 & \textbf{56.1 $\pm$ 2.4} & 95.1 $\pm$ 0.3 \\
8 & 45.1 $\pm$ 3.7 & 48.2 $\pm$ 2.8 & 50.6 $\pm$ 2.1 & 53.3 $\pm$ 2.1 & \textbf{56.4 $\pm$ 1.8} & 95.1 $\pm$ 0.3 \\
11 & 45.1 $\pm$ 3.2 & 50.9 $\pm$ 2.1 & 51.3 $\pm$ 2.2 & 55.0 $\pm$ 1.4 & \textbf{57.3 $\pm$ 1.5} & 95.1 $\pm$ 0.3 \\
14 & 46.3 $\pm$ 0.0 & 51.9 $\pm$ 0.0 & 52.3 $\pm$ 0.0 & 54.4 $\pm$ 0.0 & \textbf{57.1 $\pm$ 0.0} & 95.1 $\pm$ 0.3 \\
\hline
Avg & 44.7 $\pm$ 3.7 & 48.3 $\pm$ 3.0 & 50.2 $\pm$ 2.1 & 52.5 $\pm$ 2.1 & \textbf{55.5 $\pm$ 1.9} & 95.1 $\pm$ 0.3 \\
\bottomrule
\end{tabular}

  }
  \end{scriptsize}
\end{table}

\begin{table}
  \centering
  \caption{Cross-Person Older Adults (Accuracy)}
  \label{table:person_older_acc}
  \begin{scriptsize}
  {\renewcommand{\arraystretch}{1.4}
  \begin{tabular}{ccccccc}
\toprule
$n$ & No Adaptation & CoDATS & CoDATS-WS & \textit{CALDA} & \textit{CALDA-WS} & Train/Target \\
\midrule
2 & 22.1 $\pm$ 5.3 & 21.9 $\pm$ 3.8 & 27.6 $\pm$ 2.8 & 26.2 $\pm$ 6.0 & \textbf{35.8 $\pm$ 3.2} & 91.7 $\pm$ 0.6 \\
7 & 20.8 $\pm$ 4.5 & 21.8 $\pm$ 3.6 & 26.4 $\pm$ 2.7 & 27.1 $\pm$ 5.5 & \textbf{35.9 $\pm$ 5.3} & 91.7 $\pm$ 0.6 \\
12 & 23.4 $\pm$ 4.5 & 20.6 $\pm$ 2.9 & 26.8 $\pm$ 3.5 & 28.5 $\pm$ 5.4 & \textbf{37.3 $\pm$ 4.9} & 91.7 $\pm$ 0.6 \\
17 & 23.3 $\pm$ 2.7 & 20.4 $\pm$ 2.1 & 27.4 $\pm$ 2.6 & 29.3 $\pm$ 3.0 & \textbf{38.4 $\pm$ 2.7} & 91.7 $\pm$ 0.6 \\
22 & 22.9 $\pm$ 1.7 & 21.7 $\pm$ 2.2 & 28.5 $\pm$ 2.7 & 28.4 $\pm$ 1.9 & \textbf{38.5 $\pm$ 3.2} & 91.7 $\pm$ 0.6 \\
\hline
Avg & 22.5 $\pm$ 3.7 & 21.3 $\pm$ 2.9 & 27.3 $\pm$ 2.9 & 27.9 $\pm$ 4.4 & \textbf{37.2 $\pm$ 3.9} & 91.7 $\pm$ 0.6 \\
\bottomrule
\end{tabular}

  }
  \end{scriptsize}
\end{table}

\begin{table}
  \centering
  \caption{Cross-Time Younger Adults (Accuracy)}
  \label{table:time_younger_acc}
  \begin{scriptsize}
  {\renewcommand{\arraystretch}{1.4}
  \begin{tabular}{cccccc}
\toprule
No Adaptation & CoDATS & CoDATS-WS & \textit{CALDA} & \textit{CALDA-WS} & Train/Target \\
\midrule
{\em week 0} $\rightarrow$ 4 & & & & & \\
61.7 $\pm$ 5.7 & 61.3 $\pm$ 7.7 & 65.1 $\pm$ 9.7 & 63.1 $\pm$ 9.2 & \textbf{67.3 $\pm$ 10.9} & 97.1 $\pm$ 0.9 \\
{\em week 4 $\rightarrow$ 0} & & & & & \\
60.1 $\pm$ 7.8 & 56.7 $\pm$ 8.9 & 56.8 $\pm$ 7.9 & \textbf{61.9 $\pm$ 10.4} & 60.1 $\pm$ 8.6 & 95.2 $\pm$ 1.3 \\
\hline
{\em Avg} & & & & & \\
60.9 $\pm$ 6.7 & 59.0 $\pm$ 8.3 & 61.0 $\pm$ 8.8 & 62.5 $\pm$ 9.8 & \textbf{63.7 $\pm$ 9.7} & 96.2 $\pm$ 1.1 \\
\bottomrule
\end{tabular}

  }
  \end{scriptsize}
\end{table}

\begin{table}
  \centering
  \caption{Cross-Time Older Adults (Accuracy)}
  \label{table:time_older_acc}
  \begin{scriptsize}
  {\renewcommand{\arraystretch}{1.4}
  \begin{tabular}{cccccc}
\toprule
No Adaptation & CoDATS & CoDATS-WS & \textit{CALDA} & \textit{CALDA-WS} & Train/Target \\
\midrule
{\em week 0 $\rightarrow$ 1} & & & & & \\
27.3 $\pm$ 11.4 & 27.3 $\pm$ 12.7 & 35.5 $\pm$ 11.1 & 34.2 $\pm$ 11.4 & \textbf{40.9 $\pm$ 15.6} & 96.2 $\pm$ 2.2 \\
{\em week 0 $\rightarrow$ 2} & & & & & \\
25.6 $\pm$ 13.3 & 28.6 $\pm$ 12.6 & 33.4 $\pm$ 13.7 & 32.5 $\pm$ 12.0 & \textbf{37.8 $\pm$ 16.7} & 96.1 $\pm$ 2.0 \\
{\em week 0 $\rightarrow$ 3} & & & & & \\
27.0 $\pm$ 16.2 & 25.9 $\pm$ 15.4 & 34.2 $\pm$ 15.1 & 28.1 $\pm$ 16.1 & \textbf{34.4 $\pm$ 18.6} & 96.1 $\pm$ 2.0 \\
{\em week 1 $\rightarrow$ 0} & & & & & \\
28.8 $\pm$ 14.8 & 29.2 $\pm$ 11.5 & \textbf{36.8 $\pm$ 13.7} & 31.8 $\pm$ 13.2 & 36.1 $\pm$ 16.0 & 94.6 $\pm$ 3.0 \\
{\em week 2 $\rightarrow$ 0} & & & & & \\
27.4 $\pm$ 17.4 & 28.8 $\pm$ 14.1 & 36.3 $\pm$ 18.8 & 29.2 $\pm$ 12.6 & \textbf{36.5 $\pm$ 16.9} & 94.6 $\pm$ 3.0 \\
{\em week 3 $\rightarrow$ 0} & & & & & \\
25.2 $\pm$ 15.7 & 27.9 $\pm$ 13.3 & 32.0 $\pm$ 14.9 & 27.7 $\pm$ 14.0 & \textbf{34.4 $\pm$ 15.4} & 94.6 $\pm$ 3.0 \\
\hline
{\em Avg} & & & & & \\
26.9 $\pm$ 14.8 & 27.9 $\pm$ 13.3 & 34.7 $\pm$ 14.5 & 30.6 $\pm$ 13.2 & \textbf{36.7 $\pm$ 16.5} & 95.4 $\pm$ 2.5 \\
\bottomrule
\end{tabular}

  }
  \end{scriptsize}
\end{table}

In the main paper we include results evaluated in terms of AUC due to the imbalanced nature of the datasets. However, prior works often use accuracy as the evaluation metric. Here we additionally include the results measured in terms of accuracy, shown in Tables~\ref{table:person_younger_acc} through~\ref{table:time_older_acc}.

\begin{table}
  \centering
  \caption{Cross-Person Younger Adults (AUC) - Mapped}
  \label{table:person_younger_mapped}
  \begin{scriptsize}
  {\renewcommand{\arraystretch}{1.4}
  \begin{tabular}{ccccccc}
\toprule
$n$ & None & CoDATS & CoDATS-WS & \textit{CALDA} & \textit{CALDA-WS} \\
\midrule
2 & 0.686 $\pm$ 0.053 & 0.706 $\pm$ 0.033 & 0.729 $\pm$ 0.027 & 0.750 $\pm$ 0.038 & \textbf{0.785 $\pm$ 0.028} \\
5 & 0.707 $\pm$ 0.037 & 0.742 $\pm$ 0.030 & 0.751 $\pm$ 0.025 & 0.795 $\pm$ 0.019 & \textbf{0.829 $\pm$ 0.010} \\
8 & 0.713 $\pm$ 0.037 & 0.762 $\pm$ 0.017 & 0.764 $\pm$ 0.019 & 0.805 $\pm$ 0.017 & \textbf{0.836 $\pm$ 0.010} \\
11 & 0.731 $\pm$ 0.022 & 0.781 $\pm$ 0.018 & 0.782 $\pm$ 0.012 & 0.813 $\pm$ 0.011 & \textbf{0.841 $\pm$ 0.008} \\
14 & 0.730 $\pm$ 0.000 & 0.792 $\pm$ 0.000 & 0.781 $\pm$ 0.000 & 0.820 $\pm$ 0.000 & \textbf{0.842 $\pm$ 0.000} \\
\hline
Avg & 0.713 $\pm$ 0.030 & 0.757 $\pm$ 0.020 & 0.761 $\pm$ 0.017 & 0.797 $\pm$ 0.017 & \textbf{0.827 $\pm$ 0.011} \\
\bottomrule
\end{tabular}

  }
  \end{scriptsize}
\end{table}

\begin{table}
  \centering
  \caption{Cross-Person Older Adults (AUC) - Mapped}
  \label{table:person_older_mapped}
  \begin{scriptsize}
  {\renewcommand{\arraystretch}{1.4}
  \begin{tabular}{ccccccc}
\toprule
$n$ & None & CoDATS & CoDATS-WS & \textit{CALDA} & \textit{CALDA-WS} \\
\midrule
2 & 0.643 $\pm$ 0.051 & 0.602 $\pm$ 0.054 & 0.677 $\pm$ 0.037 & 0.695 $\pm$ 0.060 & \textbf{0.777 $\pm$ 0.041} \\
7 & 0.632 $\pm$ 0.051 & 0.611 $\pm$ 0.041 & 0.677 $\pm$ 0.034 & 0.706 $\pm$ 0.038 & \textbf{0.794 $\pm$ 0.022} \\
12 & 0.632 $\pm$ 0.059 & 0.620 $\pm$ 0.036 & 0.684 $\pm$ 0.028 & 0.730 $\pm$ 0.039 & \textbf{0.799 $\pm$ 0.023} \\
17 & 0.637 $\pm$ 0.039 & 0.630 $\pm$ 0.025 & 0.699 $\pm$ 0.025 & 0.724 $\pm$ 0.022 & \textbf{0.803 $\pm$ 0.014} \\
22 & 0.659 $\pm$ 0.032 & 0.649 $\pm$ 0.028 & 0.707 $\pm$ 0.020 & 0.734 $\pm$ 0.018 & \textbf{0.804 $\pm$ 0.013} \\
\hline
Avg & 0.640 $\pm$ 0.046 & 0.622 $\pm$ 0.037 & 0.689 $\pm$ 0.029 & 0.718 $\pm$ 0.035 & \textbf{0.795 $\pm$ 0.022} \\
\bottomrule
\end{tabular}

  }
  \end{scriptsize}
\end{table}

\begin{table}
  \centering
  \caption{Cross-Time Younger Adults (AUC) - Mapped}
  \label{table:time_younger_mapped}
  \begin{scriptsize}
  {\renewcommand{\arraystretch}{1.4}
  \begin{tabular}{cccccc}
\toprule
None & CoDATS & CoDATS-WS & \textit{CALDA} & \textit{CALDA-WS} & Train/Target \\
\midrule
{\em week 0 $\rightarrow$ 4} & & & & & \\
0.824 $\pm$ 0.041 & 0.792 $\pm$ 0.035 & 0.838 $\pm$ 0.046 & 0.859 $\pm$ 0.065 & \textbf{0.883 $\pm$ 0.046} & 0.996 $\pm$ 0.002 \\
{\em week 4 $\rightarrow$ 0} & & & & & \\
0.801 $\pm$ 0.044 & 0.783 $\pm$ 0.083 & 0.802 $\pm$ 0.060 & 0.838 $\pm$ 0.049 & \textbf{0.842 $\pm$ 0.053} & 0.991 $\pm$ 0.003 \\
\hline
{\em Avg} & & & & & \\
0.813 $\pm$ 0.042 & 0.787 $\pm$ 0.059 & 0.820 $\pm$ 0.053 & 0.849 $\pm$ 0.057 & \textbf{0.862 $\pm$ 0.050} & 0.993 $\pm$ 0.002 \\
\bottomrule
\end{tabular}

  }
  \end{scriptsize}
\end{table}

\begin{table}
  \centering
  \caption{Cross-Time Older Adults (AUC) - Mapped}
  \label{table:time_older_mapped}
  \begin{scriptsize}
  {\renewcommand{\arraystretch}{1.4}
  \begin{tabular}{cccccc}
\toprule
None & CoDATS & CoDATS-WS & \textit{CALDA} & \textit{CALDA-WS} & Train/Target \\
\midrule
{\em week 0 $\rightarrow$ 1} & & & & & \\
0.632 $\pm$ 0.142 & 0.645 $\pm$ 0.132 & 0.734 $\pm$ 0.141 & 0.703 $\pm$ 0.114 & \textbf{0.769 $\pm$ 0.143} & 0.993 $\pm$ 0.009 \\
{\em week 0 $\rightarrow$ 2} & & & & & \\
0.719 $\pm$ 0.136 & 0.660 $\pm$ 0.137 & 0.727 $\pm$ 0.137 & 0.722 $\pm$ 0.125 & \textbf{0.802 $\pm$ 0.129} & 0.993 $\pm$ 0.007 \\
{\em week 0 $\rightarrow$ 3} & & & & & \\
0.656 $\pm$ 0.156 & 0.590 $\pm$ 0.169 & 0.692 $\pm$ 0.144 & 0.670 $\pm$ 0.199 & \textbf{0.766 $\pm$ 0.129} & 0.990 $\pm$ 0.010 \\
{\em week 1 $\rightarrow$ 0} & & & & & \\
0.643 $\pm$ 0.121 & 0.625 $\pm$ 0.122 & 0.699 $\pm$ 0.120 & 0.661 $\pm$ 0.123 & \textbf{0.756 $\pm$ 0.124} & 0.985 $\pm$ 0.014 \\
{\em week 2 $\rightarrow$ 0} & & & & & \\
0.665 $\pm$ 0.134 & 0.648 $\pm$ 0.141 & 0.704 $\pm$ 0.138 & 0.678 $\pm$ 0.141 & \textbf{0.709 $\pm$ 0.116} & 0.985 $\pm$ 0.014 \\
{\em week 3 $\rightarrow$ 0} & & & & & \\
0.656 $\pm$ 0.156 & 0.629 $\pm$ 0.144 & 0.658 $\pm$ 0.132 & 0.607 $\pm$ 0.211 & \textbf{0.741 $\pm$ 0.117} & 0.985 $\pm$ 0.014 \\
\hline
{\em Avg} & & & & & \\
0.662 $\pm$ 0.141 & 0.633 $\pm$ 0.141 & 0.702 $\pm$ 0.136 & 0.673 $\pm$ 0.152 & \textbf{0.757 $\pm$ 0.127} & 0.989 $\pm$ 0.011 \\
\bottomrule
\end{tabular}

  }
  \end{scriptsize}
\end{table}

Because CASAS MAR/YA and CASAS MAR/OA have different label sets, we performed experiments to validate that this is not the cause of the increased difficulty of the CASAS MAR/OA dataset. We mapped the \textit{cook} and \textit{eat} labels for CASAS MAR/YA to \textit{mealtime} and removed the \textit{errands}, \textit{hobby}, \textit{relax}, and \textit{sleep} labels from CASAS MAR/OA. This yields both datasets consisting of the same 5 labels: \textit{work}, \textit{travel}, \textit{exercise}, \textit{hygiene}, and \textit{mealtime}. The results are shown in Tables~\ref{table:person_younger_mapped} through \ref{table:time_older_mapped}. The train on target performance for cross-person younger adults is 0.992 $\pm$ 0.001 and for older adults is 0.981 $\pm$ 0.002. In Table~\ref{table:time_older_mapped}, we also exclude data from one participant due to there not being sufficient data for train/validation/test sets after mapping. As expected, model performance on the older adult dataset is lower than on the younger adult dataset, agreeing with the results in the main paper.

\end{document}